\documentclass{article}

\usepackage{arxiv}

\usepackage[utf8]{inputenc} % allow utf-8 input
\usepackage[T1]{fontenc}    % use 8-bit T1 fonts
\usepackage{hyperref}       % hyperlinks
\usepackage{url}            % simple URL typesetting
\usepackage{booktabs}       % professional-quality tables
\usepackage{amsfonts}       % blackboard math symbols
\usepackage{nicefrac}       % compact symbols for 1/2, etc.
\usepackage{microtype}      % microtypography
\usepackage{lipsum}		% Can be removed after putting your text content
\usepackage{graphicx}
\usepackage{amsmath}
\usepackage{lscape}
\usepackage{natbib}
\usepackage{doi}

\newcommand{\indep}{\rotatebox[origin=c]{90}{$\models$}}

\title{How should we proxy for race/ethnicity? Comparing Bayesian Improved Surname Geocoding to Machine Learning methods}

\author{ \href{https://orcid.org/0000-0000-0000-0000}{\includegraphics[scale=0.06]{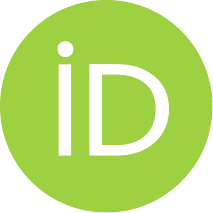}\hspace{1mm}Ari Decter-Frain}\\%\thanks{} \\
	Jeb E. Brooks School of Public Policy\\
	Cornell University\\
	Ithaca, NY, USA \\
	\texttt{agd75@cornell.edu} \\
}

\renewcommand{\shorttitle}{Benchmarking BISG}

\hypersetup{
pdftitle={A template for the arxiv style},
pdfsubject={q-bio.NC, q-bio.QM},
pdfauthor={David S.~Hippocampus, Elias D.~Striatum},
pdfkeywords={First keyword, Second keyword, More},
}

\begin{document}
\maketitle

\begin{abstract}
	Bayesian Improved Surname Geocoding (BISG) is the most popular method for proxying race/ethnicity in voter registration files that do not contain it. This paper benchmarks BISG against a range of previously untested machine learning alternatives, using voter files with self-reported race/ethnicity from California, Florida, North Carolina, and Georgia. This analysis yields three key findings. First, machine learning consistently outperforms BISG at individual classification of race/ethnicity. Second, BISG and machine learning methods exhibit divergent biases for estimating regional racial composition. Third, the performance of all methods varies substantially across states. These results suggest that pre-trained machine learning models are preferable to BISG for individual classification. Furthermore, mixed results across states underscore the need for researchers to empirically validate their chosen race/ethnicity proxy in their populations of interest.
\end{abstract}

% keywords can be removed

\section{Introduction}

Political science research often requires constructing a race/ethnicity proxy variable for datasets that do not contain it, like voter registration files, lists of electoral candidates, or political donation records. Constructing such a proxy is an important step for conducting ecological inference in voting rights litigation (\cite{barreto_estimating_2019}, \cite{imai_improving_2016}), redistricting (\cite{deluca_letter_nodate}, \cite{kenny_use_2021}), and substantive research on the role of race/ethnicity in politics (\cite{enos_what_2016}, \cite{enos_can_2019},  \cite{grumbach_race_2020}). The most common method for proxying race/ethnicity is Bayesian Improved Surname Geocoding (BISG), which uses Bayes' rule to compute a probability distribution over race/ethnicity categories conditional on a voter's surname and where they live (\cite{elliott_new_2008, elliott_using_2009}). BISG has attained widespread popularity due to its parsimony, computational efficiency, and superior performance when compared to existing alternatives, namely spatial interpolation of Census racial-ethnic composition from Census geographies (\cite{imai_improving_2016}, \cite{clark_minmaxing_2021}, \cite{shah_comparing_2017}). 

While BISG performs well compared to the small suite of existing alternatives, it has not yet been benchmarked against machine learning (ML) models, which can produce race/ethnicity predictions from more flexible and potentially more accurate models. In this paper I present the results of such a benchmark. I train a range of machine learning models using voter registration data from Florida, Georgia, North Carolina, and a portion of California where voters self-report their race/ethnicity upon registration. The registries in these states contain over 26 million labelled observations, which equates to greater than a five percent non-representative sample of the United States electorate. I then compare BISG against predictions from these models made out-of-state. 

I find that machine learning models consistently outperform BISG at individual classification, and perform roughly as well at estimating precinct composition. Furthermore, the performance of all methods varies substantially across states and neighborhoods. This underscores the need for researchers to empirically validate their use of \textit{any} race/ethnicity proxying method for each new application. I also discuss the implications of these results for existing and future research using race/ethnicity proxies. %\footnote{Raw voter data used in this paper cannot be shared directly. An R package enabling researchers to produce race/ethnicity predictions from pre-trained machine learning models is in development and will be available should the article be accepted for publication}.

%Machine learning models pre-trained on data from all four states are accessible via the R package \textit{estodemo}, which is currently accessible on github. Researchers can install this package and input their own data to get predictions based on any of the models described in this paper, with roughly the same computational burden as standard BISG.

\section{Methods for imputing race/ethnicity}
\label{sec:impute_methods}

\subsection*{BISG and extensions}

BISG computes the probability of race given a voter's surname and geographic location, $P(R=r|S=s,G=g)$, using Bayes theorem. Assuming $P(G|R) \indep P(S|R)$,

\begin{equation}
        P(R=r|S=s,G=g) \propto P(G=g|R=r)P(R=r|S=s)    
        \label{eq:1}
\end{equation}

The probability $P(G=g|R=r)$ can be obtained from Census summary tables by taking the number of people of race/ethnicity $R = r$ in neighborhood $G = g$ divided by the total number of people of race/ethnicity $r$. Recent research has shown that the imputations become more accurate as the geographic areas used to define $G$ becomes smaller. Most performance gains are achieved by matching people to geographic areas by zip code, although the best results are still obtained by geocoding the data and matching voters to blocks (\cite{clark_minmaxing_2021}). In this paper I use block-level geographic information from 2020 Decennial Census throughout. 

The probability of race given surname, $P(R=r|S=s)$, comes directly from the Census Bureau's surname lists which contain the proportion of all Decennial Census respondents with each surname in each racial-ethnic category. I use the \texttt{merge\_surnames} function from the \textit{wru} package to match voters with the correct $P(R|S)$ (\cite{wru}). I then compute posterior probability $P(R|G,S)$ by multiplying the two quantities together for each race category and dividing by the sum across all race categories to normalize. For comparison, I also use the \textit{wru} package to perform BISG, and find that the results of my calculations match closely to the package (see appendices A and B).

Previous work to extend BISG has involved layering on additional assumptions that enable the inclusion of more information within the framework of the Bayes' rule formula. Given a series of conditional independence assumptions, various extensions can be made by altering equation 1,

\begin{equation}
    P(R=r|S=s,G=g,X=x,Y=y,...) \propto \\ P(G=g|R=r)(X=x|R=r)(Y=y|R=r)...P(R=r|S=s)   
    \label{eq:2}
\end{equation}

Where $X, Y$, and so on are any voter characteristics whose probability conditional on race is computed from available data. Previous work has taken this approach to add first names (\cite{voicu_using_2018}), age, gender, and party affiliation\footnote{\cite{imai_improving_2016} make a slightly different assumption that is weaker and adds the need for an expectation-maximization step to the computation of the posterior distribution. They note that in their empirical investigations the choice of their approach over the one described here does not make a major difference for performance} (\cite{imai_improving_2016}), each yielding better performance than when using surname and geography alone. 

Here, in addition to BISG, I also consider an extended method that combines information about geographic, surname, middle name and first name using \ref{eq:2}. For each, I construct four separate lookup tables containing their probability given race-ethnicity. Each of the four tables uses data from three of the four states. Then, to compute estimates in any given state, I use the probabilities from the lookup table that does not contain that state. Therefore, extensions of BISG make use of data from out-of-state.

\subsection*{Machine learning methods}

In equations \ref{eq:1} and \ref{eq:2} every term used to generate predictions is known and no parameters are estimated prior to computing a prediction. This is an advantage of BISG -- predictions follow directly from an application of Bayes Theorem and so require no information beyond indirect data from the Census Bureau. However, data do exist with which to estimate unknown parameters in a more flexible model. This fact motivates experimenting with alternative ways of predicting the probability distribution of race/ethnicity. 

Here I consider what has become a standard suite of machine learning methods geared toward prediction problems \citep{hastie01statisticallearning}. First, I fit straightforward multinomial regression models, where the probabilistic inputs to BISG are instead summed together to generate predictions using weights learned from observed data. Second, I fit a multinomial regression adding an elasticnet penality term, which shrinks the learned weights in the model towards zero to reduce the risk of overfitting. Third, I fit random forests. Random forests average over multiple decision trees, each of which partitions subsets of observations into classification groups. Fourth, I fit gradient boosted decision trees, which greedily build random forests such that each subsequent tree minimizes the remaining loss from the previous combination of trees. These represent four of the most standard alternatives for multinomial classification tasks where prediction is the primary objective.

\section{Data and methods}

Using ML to proxy race/ethnicity carries a major risk of over-fitting to the characteristics of states for which labelled data is available. Existing research on ML for race-ethnicity classification has achieved exceptional performance when models are trained in the same geographic regions where they are tested (\cite{sood_predicting_2018}, \cite{lee_name_2017}), but political scientists and practitioners do not need to impute the race/ethnicity of voters in the regions where race/ethnicity is already self-reported -- they need to impute it in all other geographic regions of the United States. Furthermore, scholars have grown increasingly wary of machine learning applications across fields that show promising results but may later fail to replicate due to small methodological errors or slight changes in conditions \citep{kapoor_leakage_2022}.

To ensure a fair comparison between BISG and ML while accounting for the risk of over-fitting, I apply a maximally conservative `out-of-state' evaluation strategy. All supervised models are trained using data from three states and used to generate predictions in the fourth. Thus, I apply these models in a different context with a different cultural and institutional history than where they were trained. 

The potential for error due to over-fitting in this setting is further amplified by the relatively extreme racial and ethnic diversity of these four states. Georgia and North Carolina have among the largest Black population shares in the country, while California has the largest Asian share\footnote{With the exception of Hawaii. Typically, people self-reporting in the Census category Native Hawaiian/Pacific Islanders are included under ``Asian" (\cite{imai_improving_2016})}. Furthermore, although Florida and California are among the states with the largest Hispanic population share, their Hispanic populations are quite different. The largest inflows of Hispanics to Florida come from Cuba (41 percent), while Californian Hispanics are predominantly of Mexican origin (84 percent)\footnote{https://www.pewresearch.org/hispanic/2004/03/19/latinos-in-california-texas-new-york-florida-and-new-jersey/}. To the extent that surname frequency differs across nationalities, this could pose challenges for models fit using surnames to predict race/ethnicity. 

The data used for this analysis are voter registration records from Florida, Georgia, North Carolina, and California. The Florida and Georgia data were collected after the 2018 national election, North Carolina after the 2016 national election, and California after the 2020 election. For Florida, North Carolina, and Georgia, the data contain all registered voters and had already been geocoded as part of previous projects \footnote{The author received all data from Dr. Matt Barreto and Dr. Loren Collingwood. The data had previously been geocoded for use in consulting and other research projects, including \cite{barreto_estimating_2019} and \cite{decter-frain_comparing_2022}}. For California, only roughly twenty percent of the voter file contains self-reported race-ethnicity. These data were obtained in raw format and roughly eighty percent of them were successfully geocoded using the Census Bureau's geocoding API. The incompleteness of the California data should not meaningfully impact the conclusions in this paper, since my focus is on comparing methods for race-ethnicity imputation rather than downstream applications.

\begin{table}[h]
\centering
\begin{tabular}{lcccc}
\hline
              & Florida   & Georgia   & N. Carolina    & California  \\ \hline
N. obs        & 8,008,186 & 6,302,094 & 6,347,235      & 5,289,420\\ 
              &           &           &                &      \\
Prp. White    & .687      & .527      & .684           & .442 \\
Prp. Black    & .134      & .301      & .209           & .052 \\
Prp. Hispanic & .139      & .034      & .028           & .232 \\
Prp. Asian    & .017      & .024      & .013           & .159 \\
Prp. Other    & .023      & .114      & .066           & .115 \\
              &           &           &                &      \\
Mean Age      & 55.2      & 45.6      & 49.0           & 41.8 \\
              &           &           &                &      \\
Prp. Male     & .446      & .465.     & .442           & .094 \\
Prp. Female   & .541      & .533      & .525           & .083 \\ 
Prp. Unknown  & .013      & .001      & .032           & .822 \\\hline
\end{tabular}
\caption{Self-reported demographic characteristics of voters in Florida, Georgia, North Carolina, and a subset of voters from California.}
\label{tab:demogs}
\end{table}

Table \ref{tab:demogs} presents the aggregate demographic characteristics of the voters in each state after pre-processing. For all four states, the data contain voters' first, middle, and last names, along with their gender, age, and self-reported race-ethnicity. The options for self-reporting race-ethnicity in all four states are White, Black, Hispanic, Asian, Other, and Unknown. For each dataset, voters without unique IDs and who could not be successfully geocoded were removed. All voters whose race was categorized as missing or `Unknown' were also removed. I identified voters' Census blocks of residence by using the sf package in R (\cite{sf}) to conduct spatial joins to  2020 block geographic regions using their geolocation.

I trained all model types described in section 2.2. I used the tidymodels framework in R to tune, train, and evaluate the models \citep{tidymodels}. For each model type, I tuned and trained four separate models, each completely withholding data from one state. The model tuning procedure involved constructing a Latin hypercube grid of possible hyperparameter combinations and iterating over the grid using five-fold cross-validation to identify the best-performing combination. For this tuning procedure I used a random sample of the 100 thousand voters from the data. Then, using the best-performing set of hyperparameters from the tuning procedure, I trained each model on a one million-voter sample. I repeated this procedure twice using two different sets of predictors. First, I used only the ten probabilistic inputs to a basic BISG-based prediction. Second, I used an expanded set that also includes probabilistic inputs based on voters' first and middle names. Finally, I used the trained models to generate predictions for all voters in each held-out state. I present an evaluation of these models compared to BISG in the next section.

\section{Out-of-state validation}

Here I compare the performance of different race/ethnicity proxy methods at the individual and aggregate level based on an out-of-state validation exercise. I tuned and trained each model four times, each time leaving out data from one of the four states. Using each of these models I generated predictions in the fourth, held-out state, and evaluated these predictions against self-reported race/ethnicity. I then compared the performance of each method against BISG. 

The inputs to the different methods are exactly the same for all comparisons, meaning that the only difference between machine learning methods and BISG is the computations applied to those inputs to generate predictions. Furthermore, all ML models are evaluated out-of-state. For example, wherever performance is reported for an ML model in Florida, that model was trained using voter data from California, Georgia, and North Carolina, before being used to make predictions in Florida.

In the following subsections I present figures which summarize performance differences between BISG and ML models in general. For a more detailed comparison between models, appendices A and B contain tables with the specific values underlying these figures. From these appendices, two results are worth mentioning here. First, results using my implementation of BISG are similar to results when using the popular \textit{wru} package. 

Second, both BISG and ML approaches struggle with proxying the `other' race/ethnicity category. For this class, area under the curve for individual classification ranges between 0.5 and 0.7 for most models, compared to .8 to 1 for the other race/ethnicities. The `other' category may be difficult to model because it amalgamates a diverse group of individuals, including those who identify with more than one race/ethnicity. Researchers typically tolerate poor performance on the `other' category because the group represents less than one percent of the population and because substantive questions usually pertain to the more well-defined groups.

\subsection*{Classifying individual voters}

\begin{figure}[ht!]
    \centering
    \includegraphics[width=\textwidth]{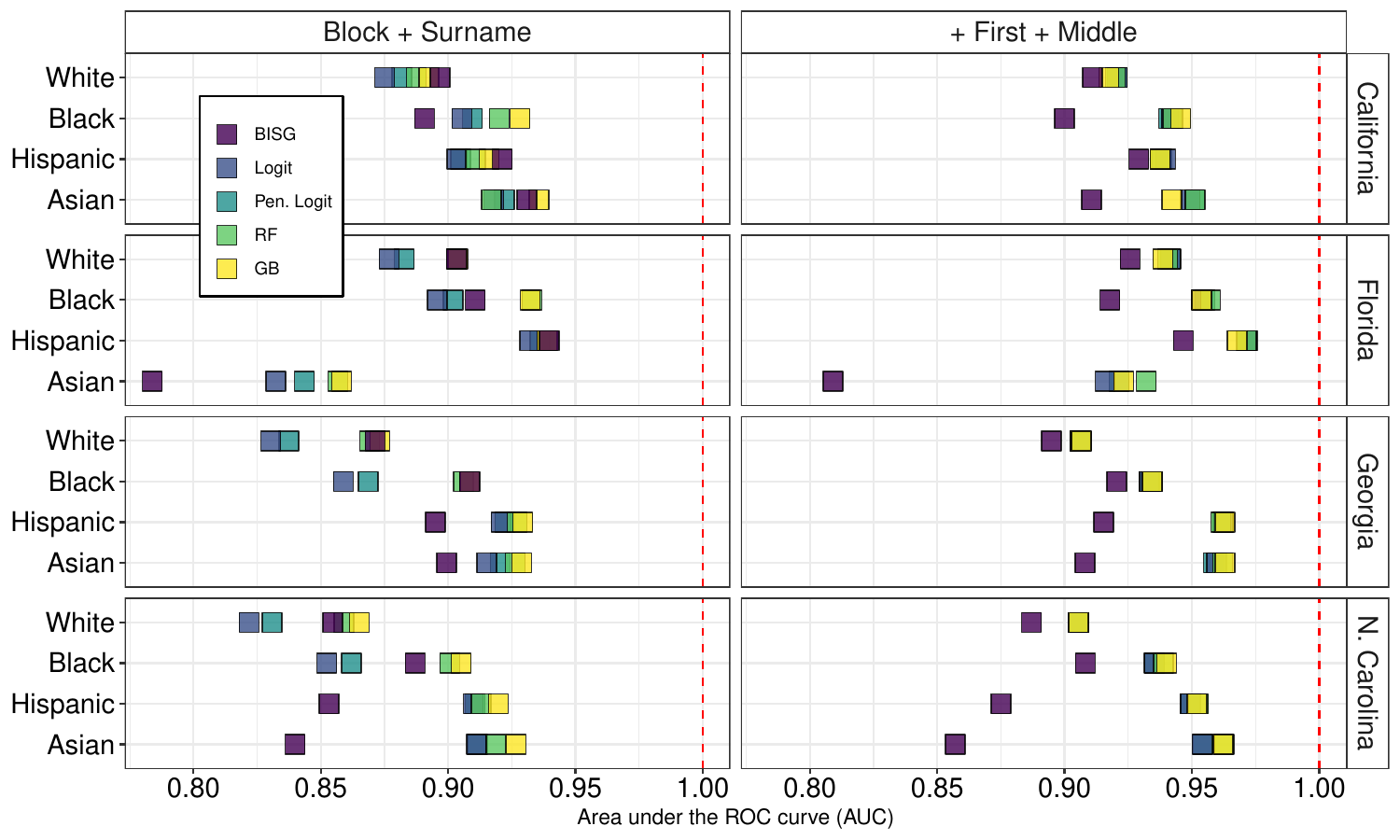}
    \caption{Comparison of area under the receiver operating curve (AUC) between BISG and ML methods. Left panels use probabilistic information about block and surname only. Right panels use these, plus information about first name and middle name. Machine learning models are trained on three states and used to generate predictions in the fourth.}
    \label{fig:class_res}
\end{figure}

I first examined the overall performance of each method at classifying voter race/ethnicity. Figure \ref{fig:class_res} presents the area under the receiver operation curve (AUC) for all models by state, race, and set of inputs. Using a minimal set of inputs ($P(G|R)$ and $P(R|S)$), BISG's performance compared to ML appears mixed, sometimes outperforming ML and sometimes not. When using the larger pool of inputs, ML models consistently outperform BISG. Notably, when using this larger set of inputs all ML methods cluster together and attain similar performance. This implies that even a simple switch from BISG to multinomial logistic regression can substantially improve individual predictions.

\begin{figure}[ht!]
    \centering
    \includegraphics[width=\textwidth]{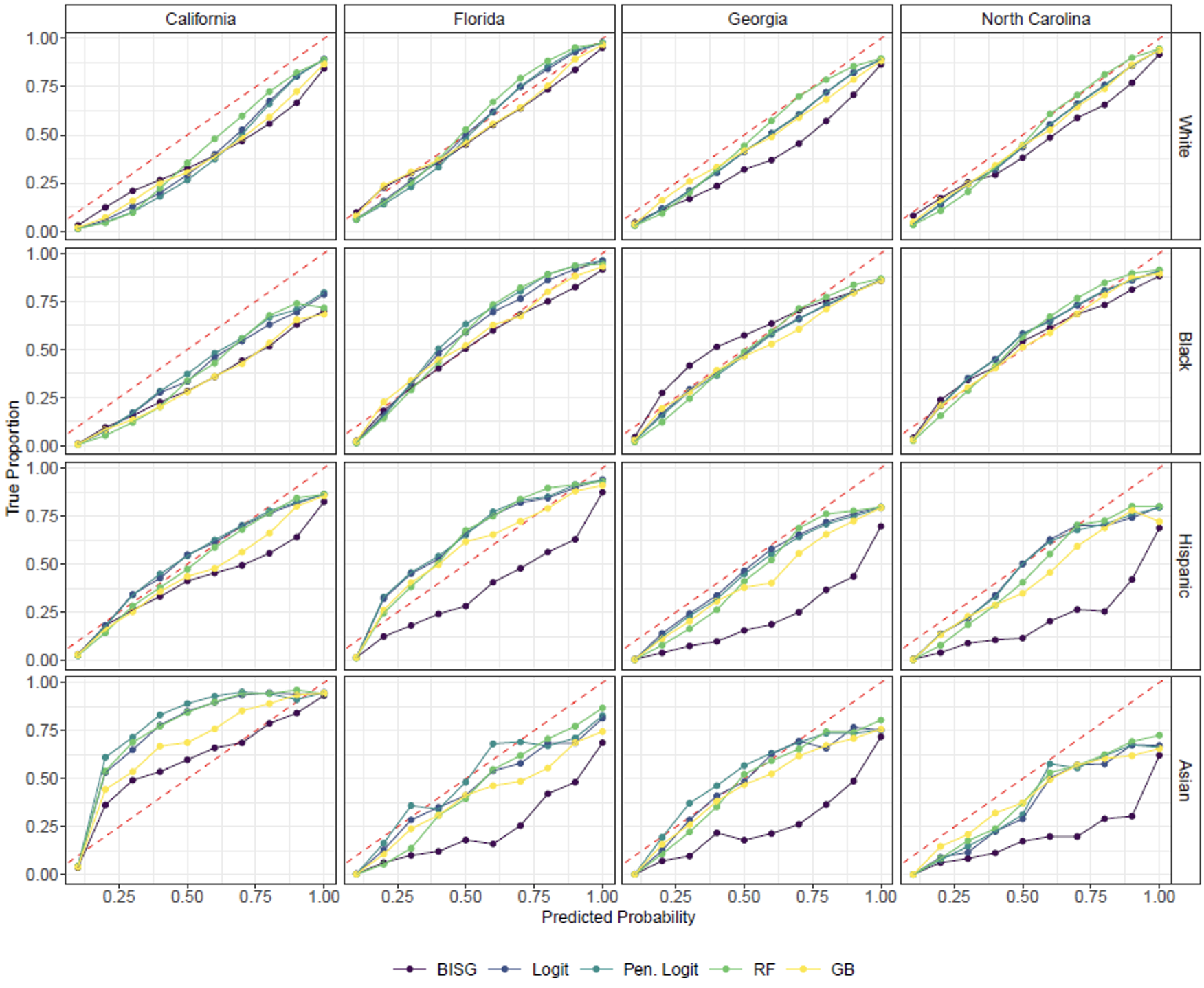}
    \caption{Calibration curves plotting the observed proportion of voters self-reporting with each race-ethnicity as a function of the predicted probabilities from each model. Both models use the full set of input data. The diagonal line indicates perfect calibration such that for any predicted probability, the actual proportion of voters with self-identifying with that race-ethnicity equals the predicted probability. All points calculated by taking the mean within bins equal to one tenth of the probability space.}
    \label{fig:calibration}
\end{figure}

Next, I examined the calibration of BISG and ML approaches. Figure \ref{fig:calibration} plots calibration curves for each method and each combination of state and race. A well-calibrated model will fall close the identity line. Being under the identity line means that too much probability is assigned to a given race/ethnicity, and being above it means not enough probability has been assigned. 

%Insufficient calibration has fairness implications \cite{corbett-davies_measure_2018}. Assigning excess probability to one race/ethnicity may lead to over-weighting that group in downstream analyses, and vice versa for under-calibration. For a given application, researchers may seek to achieve the best calibration possible, or may seek the method that minimizes over- or under-calibration for particular groups of interest.

In most cases, BISG appears equally or worse-calibrated than ML models, particularly for Asian and Hispanic voters. The one case where BISG appears better calibrated is for Asian voters in California. Both models do not assign enough probability to the Asian category for Californian voters, but the ML models appear more poorly calibrated than BISG. This poor calibration likely results from the relatively extreme size of the Asian population in California. ML models trained on other states are partially informed by the base rates of each race/ethnicity in each state, and California has the largest Asian share of any state in the country.

In general, the calibration curves for the two methods appear on the same side of the identity line. The exception is for Hispanics in Florida, where BISG is over-calibrated and the supervised method is under-calibrated. It is unclear why BISG appears over-calibrated, though the under-calibration of the supervised method is likely again due to differences in the base-rate share of Hispanics in Florida compared to the other states under study.

Taken together, these results support the conclusion that for individual prediction, more highly parameterized models leveraging observed data should be used over BISG. 

\subsection*{Estimating racial composition}

Next, I evaluated the performance of supervised methods at estimating the racial composition of a geographic area. Although many applications require analyzing voter data at the precinct level, I follow \citet{kenny_use_2021} in conducting composition evaluations for tracts. Tracts are advantageous over precincts from an analytical perspective because they are of roughly equal size and are always nested inside counties and states. Furthermore, results at the tract level should be highly consistent with those at the precinct level. 

\begin{figure}[ht!]
    \centering
    \includegraphics[width=\textwidth]{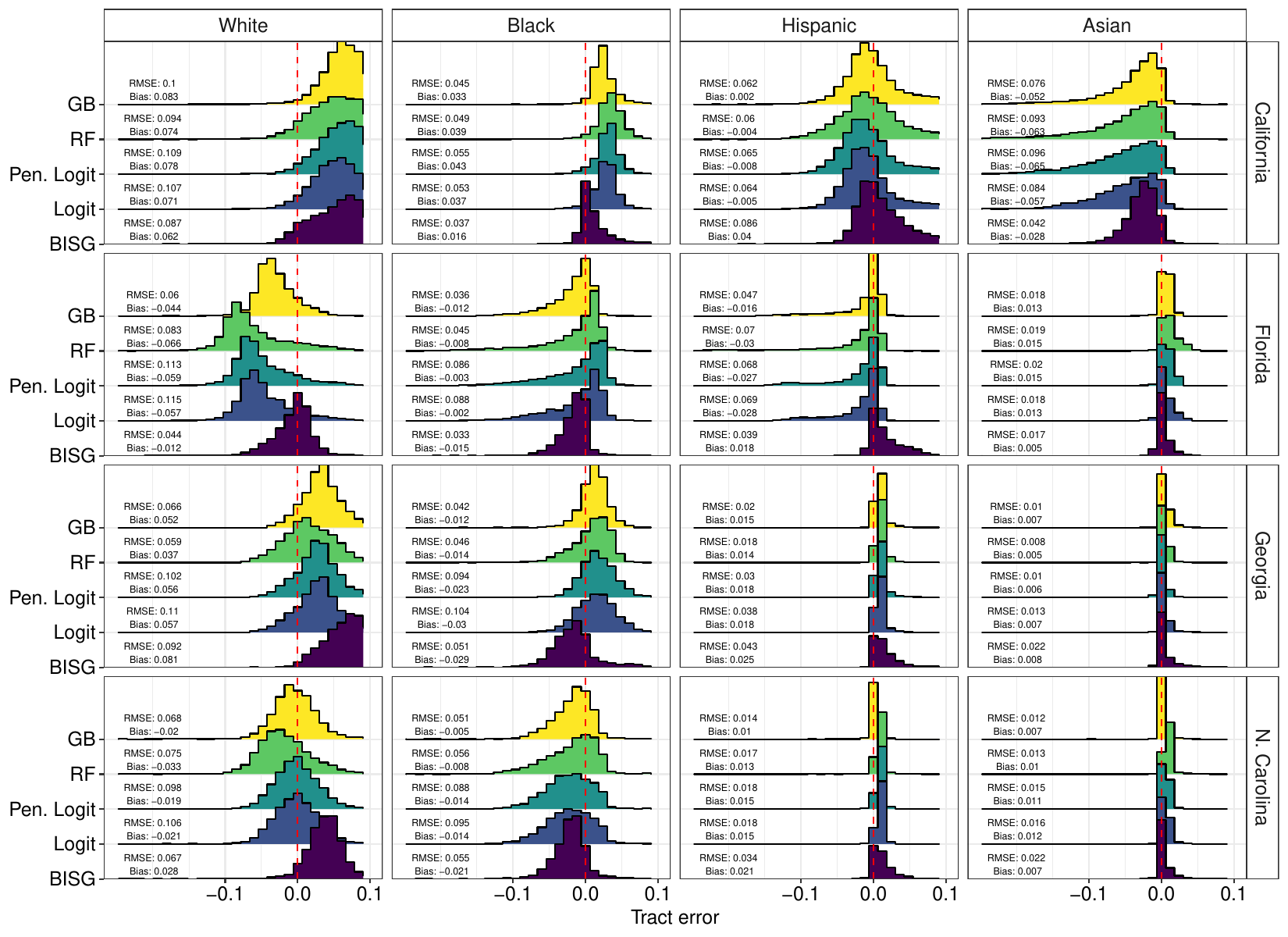}
    \caption{Density histograms of errors in estimation of tract-level racial/ethnic composition. Bins to the right of zero contain tracts for which a method has overestimated the share of voters of a given race/ethnicity, while bins to the left indicate underestimation. The RMSE and bias for each race-method-state combination are also presented.}
    \label{fig:tract_rmse}
\end{figure}

Figure \ref{fig:tract_rmse} summarizes the performance of each model for estimating tract racial/ethnic composition when using the larger set of inputs (First name, middle name, surname, block). In Georgia and North Carolina, ML models continue to outperform BISG in terms of overall error magnitude (RMSE) and bias. For Florida and California, however, results appear more mixed. BISG remains the best or one of the best methods for estimating racial composition in Florida, and in certain racial/ethnic categories in California.

Furthermore, figure \ref{fig:tract_rmse} reveals meaningful variation in the bias of all methods across states. For instance, BISG performs well at estimating the share of white voters in each precinct in Florida, the setting for the canonical BISG validation study in political science \citep{imai_improving_2016}. In every other state, however, BISG consistently overestimates the white share of precincts. Similarly for Black voters, by looking at Florida, Georgia, and North Carolina one might conclude that BISG tends to underestimate the share of black voters in each precinct. If a researcher adopted this prior belief when applying estimating the Black share of voters in California, they might end up doubly wrong; BISG overestimates the Black share of voters in California, and a researcher might assume that even these must be underestimates, given the performance of BISG in other states.

\section{Discussion}

In this paper, I conducted the first direct comparison between BISG and ML methods for constructing a race/ethnicity proxy variable in voter registration data, using data from four states and evaluating the ML models on out-of-state performance. To the author's knowledge, this is also the largest validation study for race/ethnicity proxy methods to date. The exercise has yielded three novel results. First, BISG and machine learning perform similarly for estimating aggregate racial/ethnic composition. Second, machine learning outperforms BISG at individual classification of race/ethnicity. Third, the performance of all methods varies substantially across states.

The first two points have clear implications for future research. When a researcher's goal is to estimate the racial composition of a neighborhood or precinct, the evidence presented here suggests BISG is a good tool. While it does not perform best for every race/ethnicity in all of the tested states, it performs comparably or better in many instances. For individual prediction, results are less favorable to BISG. Provided researchers have access to individuals' first and middle names, they should use ML models trained on available data from voter registries. ML consistently produced more accurate individual-level race/ethnicity proxies than BISG in every state for every race/ethnicity. Furthermore, ML methods were better-calibrated than BISG in every case except for Asians in California.

Another important finding from this benchmarking exercise is that race/ethnicity proxy methods, methods perform differently across states. For precinct composition estimation, patterns of over- or underestimation do not consistent across the country. As such, it may not be sufficient for researchers to rely on empirical validation studies in Florida \citep{imai_improving_2016} or Georgia \citep{clark_minmaxing_2021} when applying BISG in a novel context. Instead, researchers should seek to validate their use of any race/ethnicity metric in each new setting by collecting ground-truth race/ethnicity self-reports and using them for in a validation study.

Biased proxy variables can bias conclusions from downstream analyses, particularly when the bias is correlated with the outcome and predictor of interest \cite{knox_testing_2022}. Confidentiality requirements often prevent authors from publicizing raw individual-level data for research reliant on BISG. This largely prevents empirical assessments of how the results of previous research may be influenced by biases in proxied race/ethnicity. As a result, I can only speculate about the implications of the results above for existing and future research.
 
Two papers apply BISG in California, where the results above do suggest that BISG is biased. Enos, Kaufman, and Sands use BISG to construct inputs for ecological inference, to estimate changes in Black and white support for education policies as a result of the LA riots \citep{enos_can_2019}. The authors justify their use of BISG, in part, by citing the false positive rate for Black voters in Florida \cite{imai_improving_2016}. This argument does not hold given the results presented above, which show that while BISG provides fairly unbiased estimates of Black and white racial composition in Florida, it tends to overestimate both shares in California to varying degrees. The potential bias of the race/ethnicity proxy is not critical to the results of this paper since the estimated effect persists across both Black and white. However, it might perhaps cast some doubt on the relative size of the observed effects when decomposed by race/ethnicity. 

Abbott and Magazinnik also use BISG in California to study the effect of a change from at-large to ward elections on the share of Latino candidates elected to school boards \citep{abott_atlarge_2020}. BISG is used to proxy the race/ethnicity of the candidates. The authors argue that BISG is not likely biased in this setting because the independence assumption almost certainly holds. The results presented in this paper show that, empirically, BISG's predictions about Hispanic voters in California are biased. They overestimate the share of Hispanics in electoral precincts and over-assign probability to the Hispanic category in individual classification. For this application application, a pre-trained ML model would likely yield a more accurate proxy. 

The authors model the proportion seats won by Hispanics in each school board in a two-way fixed effects regression analysis where the parameter of interest interacts the Hispanic share of the school district electorate with an indicator for the presence of ward elections. The results presented above suggest that the authors' outcome of interest may be biased upward, which would not in itself affect their key estimates. However, if the bias towards overestimating Hispanic share is correlated across space with the Hispanic share of each school district, the estimated effect of switching to ward elections may also be biased. 

Additionally, the authors show that the effect of changing to ward elections on the share of Hispanic election winners is concentrated within the most segregated school districts across the state. BISG is most accurate in segregated areas \citep{decter-frain_comparing_2022}. The reported differences in the effects across districts with different levels of segregation may result from some combination of a real difference, and differences in the accuracy of BISG across segregated and diverse neighborhoods.

The above two examples illustrate the challenge of using race/ethnicity proxies in substantive research. In each case, concerns about the potential for biases to emerge from the proxy could be best addressed by manually labelling a small set of individuals and using them to compare methods and decide on the most accurate for the setting. In the latter case, the evidence presented in this paper suggests an ML model may provide more credible results than BISG. 

Beyond these examples, BISG has been used in nationwide applications \citep{brown_measurement_2021, grumbach_race_2020} and has become increasingly common in redistricting \citep{deluca_letter_nodate}, and voting rights \citep{barreto_novel_2021}. For each of these applications, BISG may or may not be the best method for constructing a proxy. The results I have presented here suggest that, in many cases, a machine learning model pre-trained on labelled observations from California, Florida, Georgia, and North Carolina may provide more accurate, better calibrated results.

\clearpage

\bibliographystyle{unsrtnat}
\bibliography{references}  %%% Uncomment this line and comment out the ``thebibliography'' section below to use the external .bib file (using bibtex) .





% \end{thebibliography}

\clearpage

\appendix

\section{Full individual-level results}

\begin{landscape}
\begin{table}
\begin{tabular}{llccccccccccccccc}
\toprule
& & \multicolumn{7}{c}{Block + Surname} & & \multicolumn{6}{c}{+ First + Middle}\\
State & Race/Eth & WRU & BISG & Logit & Logit2 & ELNET & RF & GB & & BISG & Logit & Logit2 & ELNET & RF & GB\\
\midrule
California\\
 & White & 0.893 & 0.897 & 0.875 & 0.873 & 0.882 & 0.887 & 0.892 & & 0.911 & 0.921 & 0.921 & 0.921 & 0.920 & 0.917\\
 & Black & 0.878 & 0.891 & 0.905 & 0.895 & 0.909 & 0.920 & 0.928 & & 0.900 & 0.942 & 0.945 & 0.941 & 0.943 & 0.946\\
 & Hispanic & 0.915 & 0.921 & 0.903 & 0.905 & 0.905 & 0.911 & 0.916 & & 0.929 & 0.940 & 0.941 & 0.938 & 0.937 & 0.938\\
 & Asian & 0.916 & 0.931 & 0.918 & 0.910 & 0.922 & 0.917 & 0.936 & & 0.911 & 0.949 & 0.953 & 0.951 & 0.951 & 0.942\\
 & Other & 0.527 & 0.524 & 0.508 & 0.512 & 0.513 & 0.508 & 0.512 & & 0.547 & 0.556 & 0.538 & 0.554 & 0.537 & 0.546\\
\addlinespace
Florida\\
 & White & 0.899 & 0.903 & 0.877 & 0.862 & 0.883 & 0.903 & 0.904 & & 0.926 & 0.942 & 0.942 & 0.942 & 0.941 & 0.939\\
 & Black & 0.909 & 0.911 & 0.896 & 0.877 & 0.902 & 0.933 & 0.932 & & 0.918 & 0.954 & 0.957 & 0.955 & 0.957 & 0.954\\
 & Hispanic & 0.932 & 0.940 & 0.932 & 0.930 & 0.936 & 0.939 & 0.939 & & 0.947 & 0.972 & 0.972 & 0.972 & 0.971 & 0.968\\
 & Asian & 0.778 & 0.784 & 0.832 & 0.844 & 0.844 & 0.857 & 0.858 & & 0.809 & 0.916 & 0.929 & 0.921 & 0.932 & 0.923\\
 & Other & 0.575 & 0.576 & 0.591 & 0.593 & 0.605 & 0.580 & 0.587 & & 0.591 & 0.662 & 0.646 & 0.665 & 0.644 & 0.633\\
\addlinespace
Georgia\\
 & White & 0.867 & 0.871 & 0.830 & 0.793 & 0.837 & 0.869 & 0.873 & & 0.895 & 0.906 & 0.907 & 0.906 & 0.907 & 0.907\\
 & Black & 0.904 & 0.908 & 0.859 & 0.815 & 0.869 & 0.906 & 0.909 & & 0.920 & 0.933 & 0.934 & 0.933 & 0.934 & 0.934\\
 & Hispanic & 0.873 & 0.895 & 0.921 & 0.929 & 0.922 & 0.927 & 0.929 & & 0.915 & 0.962 & 0.964 & 0.962 & 0.961 & 0.963\\
 & Asian & 0.878 & 0.899 & 0.915 & 0.927 & 0.920 & 0.926 & 0.929 & & 0.908 & 0.960 & 0.964 & 0.958 & 0.962 & 0.963\\
 & Other & 0.531 & 0.525 & 0.542 & 0.555 & 0.544 & 0.550 & 0.550 & & 0.540 & 0.566 & 0.577 & 0.561 & 0.572 & 0.569\\
\addlinespace
N. Carolina\\
& White & 0.853 & 0.855 & 0.822 & 0.790 & 0.831 & 0.859 & 0.865 & & 0.887 & 0.905 & 0.906 & 0.906 & 0.906 & 0.905\\
& Black & 0.889 & 0.887 & 0.852 & 0.821 & 0.862 & 0.901 & 0.905 & & 0.908 & 0.935 & 0.940 & 0.935 & 0.939 & 0.940\\
& Hispanic & 0.833 & 0.853 & 0.910 & 0.912 & 0.911 & 0.913 & 0.920 & & 0.875 & 0.949 & 0.952 & 0.949 & 0.953 & 0.952\\
& Asian & 0.817 & 0.840 & 0.911 & 0.915 & 0.911 & 0.919 & 0.927 & & 0.857 & 0.954 & 0.961 & 0.954 & 0.963 & 0.962\\
& Other & 0.571 & 0.555 & 0.587 & 0.594 & 0.596 & 0.595 & 0.605 & & 0.569 & 0.600 & 0.627 & 0.599 & 0.643 & 0.634\\
\bottomrule
\end{tabular}
\caption{Area under the curve (AUC) for each model for each race in each state.}
\label{aroc}
\end{table}
\end{landscape}

\section{Full precinct-level results}

\begin{landscape}
\begin{table}
\centering
\begin{tabular}{llcccccccccccccc}
\toprule 
& & \multicolumn{7}{c}{Block + Surname} & & \multicolumn{6}{c}{+ First + Middle}\\
State & Race/Eth & BISG & WRU & ELNET & Logit & Logit2 & RF & GB & & BISG & ELNET & Logit & Logit2 & RF & GB\\

\midrule
California \\
 & White & .087 & .1 & .121 & .121 & .136 & .107 & .112 & & .088 & .083 & .074 & .071 & .075 & .087\\
 & Black & .036 & .042 & .066 & .064 & .063 & .053 & .051 & & .029 & .034 & .032 & .025 & .039 & .030\\
 & Hispanic & .070 & .072 & .055 & .051 & .056 & .050 & .050 & & .064 & .048 & .049 & .045 & .045 & .047\\
 & Asian & .039 & .044 & .117 & .102 & .106 & .103 & .091 & & .041 & .084 & .073 & .067 & .094 & .067\\
 & Other & .113 & .113 & .083 & .082 & .083 & .084 & .088 & & .105 & .083 & .083 & .083 & .080 & .095\\
\addlinespace
Florida\\
 & White & .045 & .049 & .143 & .148 & .162 & .094 & .075 & & .037 & .066 & .063 & .058 & .078 & .044\\
 & Black & .038 & .046 & .101 & .106 & .125 & .039 & .035 & & .035 & .044 & .042 & .036 & .044 & .035\\
 & Hispanic & .032 & .029 & .074 & .077 & .078 & .066 & .048 & & .029 & .037 & .033 & .032 & .052 & .029\\
 & Asian & .018 &.017 & .024 & .021 & .024 & .021 & .023 & & .010 & .010 & .010 & .008 & .014 & .008\\
 & Other & .016 & .012 & .075 & .075 & .075 & .090 & .063 & & .015 & .075 & .074 & .070 & .090 & .055\\
\addlinespace
Georgia\\
 & White & .097 & .115 & .130 & .141 & .184 & .070 & .078 & & .087 & .046 & .046 & .047 & .038 & .049\\
 & Black & .059 & .077 & .120 & .132 & .177 & .056 & .052 & & .034 & .029 & .029 & .027 & .027 & .022\\
 & Hispanic & .046 & .045 & .030 & .039 & .023 & .021 & .023 & & .026 & .015 & .012 & .009 & .012 & .012\\
 & Asian & .025 & .024 & .012 & .014 & .013 & .009 & .011 & & .017 & .006 & .010 & .008 & .006 & .008\\
 & Other & .092 & .092 & .067 & .065 & .063 & .056 & .068 & & .092 & .070 & .068 & .064 & .051 & .073\\
\addlinespace
N. Carolina
 & White & .044 & .057 & .109 & .120 & .149 & .069 & .058 & & .051 & .038 & .039 & .041 & .044 & .037\\
 & Black & .041 & .051 & .093 & .104 & .129 & .038 & .033 & & .034 & .042 & .041 & .033 & .040 & .032\\
 & Hispanic & .039 & .035 & .022 & .022 & .020 & .019 & .017 & & .017 & .010 & .010 & .006 & .010 & .006\\
 & Asian & .024 & .023 & .016 & .017 & .015 & .013 & .012 & & .014 & .009 & .011 & .007 & .010 & .006\\
 & Other & .046 & .049 & .050 & .052 & .051 & .050 & .047 & & .044 & .051 & .050 & .044 & .049 & .043\\
\bottomrule
\end{tabular}
\caption{RMSE values for every model-race-state combination. All calculations weight by tract size.}
\label{tab:armse}
\end{table}
\end{landscape}

\begin{landscape}
\begin{table}
\begin{tabular}{llcccccccccccccc}
\toprule
& & \multicolumn{7}{c}{Block + Surname} & & \multicolumn{6}{c}{+ First + Middle}\\
State & Race/Eth & BISG & WRU & ELNET & Logit & Logit2 & RF & GB & & BISG & ELNET & Logit & Logit2 & RF & GB\\
\midrule
California\\
 & White & .070 & .079 & .085 & .079 & .093 & .095 & .100 & & .075 & .072 & .062 & .059 & .061 & .078\\
 & Black & .018 & .016 & .058 & .050 & .043 & .042 & .039 & & .013 & .030 & .028 & .021 & .035 & .026\\
 & Hispanic & .031 & .027 & -.015 & -.012 & -.016 & -.014 & -.012 & & .025 & -.002 & .000 & .004 & .000 & .007\\
 & Asian & -.029 & -.032 & -.082 & -.072 & -.075 & -.076 & -.070 & & -.032 & -.065 & -.057 & -.048 & -.066 & -.047\\
 & Other & -.091 & -.090 & -.046 & -.045 & -.045 & -.047 & -.057 & & -.081 & -.036 & -.033 & -.035 & -.031 & -.064\\
\addlinespace
Florida\\
 & White & -.013 & .002 & -.084 & -.081 & -.081 & -.079 & -.061 & & -.005 & -.051 & -.049 & -.046 & -.064 & -.032\\
 & Black & -.013 & -.018 & .014 & .019 & .014 & .002 & -.004 & & -.015 & -.008 & -.009 & -.012 & -.011 & -.016\\
 & Hispanic & .014 & .008 & -.025 & -.027 & -.026 & -.026 & -.015 & & .015 & -.017 & -.017 & -.013 & -.021 & -.010\\
 & Asian & .009 & .009 & .021 & .018 & .020 & .018 & .020 & & .003 & .006 & .005 & .004 & .010 & .006\\
 & Other & .002 & -.001 & .073 & .072 & .073 & .086  & .060 & & .002 & .070 & .070 & .067 & .086 & .051\\
\addlinespace
Georgia\\
 & White & .086 & .100 & .082 & .085 & .078 & .056 & .068 & & .079 & .033 & .033 & .034 & .018 & .039\\
 & Black & -.041 & -.049 & -.060 & -.069 & -.058 & -.034 & -.035 & & -.016 & .017 & .014 & .013 & .009 & .012\\
 & Hispanic & .030 & .025  & .023 & .024 & .022 & .018 & .019 & & .016 & .010 & .009 & .006 & .010 & .009\\
 & Asian & .010 & .009 & .010 & .010 & .012 & .005 & .009 & & .006 & .002 & .004 & .002 & .004 & .005\\
 & Other & -.085 & -.085 & -.056 & -.051 & -.053 & -.046 & -.060 & & -.085 & -.062 & -.060 & -.055 & -.041 & -.065\\
\addlinespace
N. Carolina
 & White & .023 & .035 & -.042 & -.045 & -.045 & -.050 & -.038 & & .040 & .001 & .001 & .003 & -.016 & -.001\\
 & Black & -.025 & -.028 & -.005 & -.004 & .000 & .000 & .001 & & -.021 & -.022 & -.022 & -.018 & -.018 & -.016\\
 & Hispanic & .028 & .023 & .021 & .022 & .020 & .018 & .015 & & .012 & .009 & .009 & .005 & .009 & .004\\
 & Asian & .010 & .010 & .015 & .016 & .014 & .012 & .009 & & .004 & .007 & .007 & .004 & .008 & .004\\
 & Other & -.035 & -.040 & .011 & .012 & .011 & .021 & .013 & & -.035 & .005 & .005 & .006 & .017 & .009\\
\bottomrule
\end{tabular}
\caption{Bias values for every model-race-state combination. All calculations weight by tract size.}
\label{tab:abias}
\end{table}
\end{landscape}

\end{document}